\documentclass[lettersize,journal]{IEEEtran}
\usepackage{amsmath,amsfonts}
\usepackage{algorithmic}
\usepackage{algorithm}
\usepackage{array}
\ifCLASSOPTIONcompsoc
  \usepackage[caption=false,font=normalsize,labelfont=sf,textfont=sf]{subfig}
\else
  \usepackage[caption=false,font=footnotesize]{subfig}
\fi
\usepackage{textcomp}
\usepackage{stfloats}
\usepackage{url}
\usepackage{verbatim}
\usepackage{graphicx}
\usepackage{cite}
\usepackage{xspace} 
\usepackage{listings}
\usepackage{xcolor}
\usepackage{colortbl} 
\usepackage{multirow}
\usepackage{multicol} 
\usepackage{bm}
\usepackage{booktabs}

\newcommand{\R}{\textsuperscript{\textregistered}\xspace}
\newcommand{\TM}{\textsuperscript{\texttrademark}\xspace}
\newcommand{\AFFECT}{AFF3CT\xspace}
\newcommand{\MIPP}{MIPP\xspace}
\newcommand{\Cxx}{C++\xspace}
\newcommand{\Cxy}[1]{C++{#1}\xspace}

\usepackage{color}

\definecolor{darkWhite}{rgb}{0.96,0.96,0.96}
\definecolor{bluekeywords}{rgb}{0.13,0.13,1}
\definecolor{greencomments}{rgb}{0,0.5,0}
\definecolor{redstrings}{rgb}{0.9,0,0}

\definecolor{Comment}{RGB}{97,161,176}

\definecolor{btfGreen}{RGB}{51,160,44}
\definecolor{btfRed}{RGB}{190,60,90}

\definecolor{bleuUni}{RGB}{0, 157, 224}
\definecolor{marronUni}{RGB}{68, 58, 49}

\definecolor{bluecite}{HTML}{009DE0}

\definecolor{Paired-2}{RGB}{166,206,227}
\definecolor{Paired-1}{RGB}{31,120,180}
\definecolor{Paired-4}{RGB}{178,223,138}
\definecolor{Paired-3}{RGB}{51,160,44}
\definecolor{Paired-6}{RGB}{251,154,153}
\definecolor{Paired-5}{RGB}{227,26,28}
\definecolor{Paired-8}{RGB}{253,191,111}
\definecolor{Paired-7}{RGB}{255,127,0}
\definecolor{Paired-10}{RGB}{202,178,214}
\definecolor{Paired-9}{RGB}{106,61,154}
\definecolor{Paired-12}{RGB}{255,255,153}
\definecolor{Paired-11}{RGB}{177,89,40}
\definecolor{Accent-1}{RGB}{127,201,127}
\definecolor{Accent-2}{RGB}{190,174,212}
\definecolor{Accent-3}{RGB}{253,192,134}
\definecolor{Accent-4}{RGB}{255,255,153}
\definecolor{Accent-5}{RGB}{56,108,176}
\definecolor{Accent-6}{RGB}{240,2,127}
\definecolor{Accent-7}{RGB}{191,91,23}
\definecolor{Accent-8}{RGB}{102,102,102}
\definecolor{Spectral-1}{RGB}{158,1,66}
\definecolor{Spectral-2}{RGB}{213,62,79}
\definecolor{Spectral-3}{RGB}{244,109,67}
\definecolor{Spectral-4}{RGB}{253,174,97}
\definecolor{Spectral-5}{RGB}{254,224,139}
\definecolor{Spectral-6}{RGB}{255,255,191}
\definecolor{Spectral-7}{RGB}{230,245,152}
\definecolor{Spectral-8}{RGB}{171,221,164}
\definecolor{Spectral-9}{RGB}{102,194,165}
\definecolor{Spectral-10}{RGB}{50,136,189}
\definecolor{Spectral-11}{RGB}{94,79,162}
\definecolor{Set1-1}{RGB}{228,26,28}
\definecolor{Set1-2}{RGB}{55,126,184}
\definecolor{Set1-3}{RGB}{77,175,74}
\definecolor{Set1-4}{RGB}{152,78,163}
\definecolor{Set1-5}{RGB}{255,127,0}
\definecolor{Set1-6}{RGB}{255,255,51}
\definecolor{Set1-7}{RGB}{166,86,40}
\definecolor{Set1-8}{RGB}{247,129,191}
\definecolor{Set1-9}{RGB}{153,153,153}
\definecolor{Set2-1}{RGB}{102,194,165}
\definecolor{Set2-2}{RGB}{252,141,98}
\definecolor{Set2-3}{RGB}{141,160,203}
\definecolor{Set2-4}{RGB}{231,138,195}
\definecolor{Set2-5}{RGB}{166,216,84}
\definecolor{Set2-6}{RGB}{255,217,47}
\definecolor{Set2-7}{RGB}{229,196,148}
\definecolor{Set2-8}{RGB}{179,179,179}
\definecolor{Dark2-1}{RGB}{27,158,119}
\definecolor{Dark2-2}{RGB}{217,95,2}
\definecolor{Dark2-3}{RGB}{117,112,179}
\definecolor{Dark2-4}{RGB}{231,41,138}
\definecolor{Dark2-5}{RGB}{102,166,30}
\definecolor{Dark2-6}{RGB}{230,171,2}
\definecolor{Dark2-7}{RGB}{166,118,29}
\definecolor{Dark2-8}{RGB}{102,102,102}
\definecolor{Reds-1}{RGB}{255,245,240}
\definecolor{Reds-2}{RGB}{254,224,210}
\definecolor{Reds-3}{RGB}{252,187,161}
\definecolor{Reds-4}{RGB}{252,146,114}
\definecolor{Reds-5}{RGB}{251,106,74}
\definecolor{Reds-6}{RGB}{239,59,44}
\definecolor{Reds-7}{RGB}{203,24,29}
\definecolor{Reds-8}{RGB}{165,15,21}
\definecolor{Reds-9}{RGB}{103,0,13}
\definecolor{Greens-1}{RGB}{247,252,245}
\definecolor{Greens-2}{RGB}{229,245,224}
\definecolor{Greens-3}{RGB}{199,233,192}
\definecolor{Greens-4}{RGB}{161,217,155}
\definecolor{Greens-5}{RGB}{116,196,118}
\definecolor{Greens-6}{RGB}{65,171,93}
\definecolor{Greens-7}{RGB}{35,139,69}
\definecolor{Greens-8}{RGB}{0,109,44}
\definecolor{Greens-9}{RGB}{0,68,27}
\definecolor{Blues-1}{RGB}{247,251,255}
\definecolor{Blues-2}{RGB}{222,235,247}
\definecolor{Blues-3}{RGB}{198,219,239}
\definecolor{Blues-4}{RGB}{158,202,225}
\definecolor{Blues-5}{RGB}{107,174,214}
\definecolor{Blues-6}{RGB}{66,146,198}
\definecolor{Blues-7}{RGB}{33,113,181}
\definecolor{Blues-8}{RGB}{8,81,156}
\definecolor{Blues-9}{RGB}{8,48,107}

\begin{document}

\title{A DSEL for High Throughput and Low Latency Software-Defined Radio on Multicore CPUs}

\author{\IEEEauthorblockN{Adrien Cassagne\IEEEauthorrefmark{1},
Romain Tajan\IEEEauthorrefmark{2},
Olivier Aumage\IEEEauthorrefmark{3},
Camille Leroux\IEEEauthorrefmark{2},
Denis Barthou\IEEEauthorrefmark{3} and
Christophe J\'ego\IEEEauthorrefmark{2}
}
\IEEEauthorblockA{\IEEEauthorrefmark{1}Sorbonne Université, CNRS, LIP6, F-75005 Paris, France}\\
\IEEEauthorblockA{\IEEEauthorrefmark{2}IMS Laboratory, UMR CNRS 5218, Bordeaux INP, University of Bordeaux, Talence, France}\\
\IEEEauthorblockA{\IEEEauthorrefmark{3}Inria, Bordeaux Institute of Technology, LaBRI/CNRS, Bordeaux, France}
}




\maketitle

\begin{abstract}
This article presents a new Domain Specific Embedded Language (DSEL) dedicated
to Software-Defined Radio (SDR). From a set of carefully designed components, it
enables to build efficient software digital communication systems, able to take
advantage of the parallelism of modern processor architectures, in a
straightforward and safe manner for the programmer. In particular, proposed DSEL
enables the combination of pipelining and sequence duplication techniques to
extract both temporal and spatial parallelism from digital communication
systems. We leverage the DSEL capabilities on a real use case: a fully digital
transceiver for the widely used DVB-S2 standard designed entirely in software.
Through evaluation, we show how proposed software DVB-S2 transceiver is able to
get the most from modern, high-end multicore CPU targets.
\end{abstract}


\section{Introduction}

\IEEEPARstart{D}{igital} communication systems are traditionally implemented
onto dedicated hardware (ASIC) to achieve high throughputs, low latencies and
energy efficiency. However, hardware implementations suffer from a long time to
market, are expensive and specific by
nature~\cite{Palkovic2010,Palkovic2012}. New communication standards such as the
5G are coming with large specifications and numerous possible
configurations~\cite{ETSI2018}. Connecting objects that exchange small amounts
of data at low rates will live together with 4K video streaming for mobile phone
games requiring high throughputs and low latencies~\cite{Rost2014}.

To meet such diverse specifications, transceivers have to be able to adapt
quickly to new configurations. Flexible, re-configurable and programmable
solutions are thus increasingly required, fueling a growing interest for the
Software-Defined Radio (SDR). It consists in processing both the Physical
(PHY) and Medium Access Control (MAC) layers in software~\cite{Mitola1993},
rather than in hardware. Shorter time to market, lower design costs, ability to
be updated, to support and to interroperate with new protocols are its main
advantages~\cite{Akeela2018}.

SDR can be implemented on various targets such as Field Programmable Gate Arrays
(FPGAs)~\cite{Coulton2004,Skey2006,Dutta2010,Shaik2013,Maheshwarappa2015,
Nivin2016}, Digital Signal Processors (DSPs)~\cite{Kaur2008,Karlsson2013,
Shaik2013} or General Purpose Processors (GPPs)~\cite{Yoge2012,Bang2014,
Meshram2019,Grayver2020}. Many SDR elementary blocks have been optimized for
Intel\R and ARM\R CPUs. High throughput results have been achieved on
GPUs~\cite{Xianjun2013,Li2014,LeGal2014a,Giard2016b,Keskin2017a}; latency
results are is still too high however to meet real time constraints and to
compete with CPU implementations~\cite{Sarkis2014,LeGal2015a,Cassagne2015c,
Giard2016b,Sarkis2016,Cassagne2016a,Cassagne2016b,LeGal2016,LeGal2017,
Leonardon2019,LeGal2019a}. This is mainly due to data transfers between the host
(CPUs) and the device (GPUs), and to the nature of GPU designs, which are not
optimized for latency efficiency. In this paper, we focus on the execution of
SDR on multicore general purpose CPUs, similar to the equipment of antennas or
transceivers.

Digital communication systems can be refined so that the transmitter and the
receiver parts are decomposed into several processing blocks connected in
a directed graph. This matches the dataflow
model~\cite{Dennis1980,Ackerman1982}: Blocks are filters and links between
blocks are data exchanges. Specific dataflow models such as the synchronous
dataflow~\cite{Lee1987} and the cyclo-static dataflow~\cite{Engels1994,
Bilsen1995} allow the expression of a static schedule for the
graph~\cite{Parks1995}. SDR however requires a parallel task graph between
stateful tasks and a dynamic schedule due to early exits, conditionals and loop
iterations. Maximizing throughput is the main objective, keeping latency as low
as possible is a secondary objective. This constrains the time taken by data
movements and requires optimizing for parallelism. For a SDR operating in
antennas or transceivers, memory footprint is not an issue and all tasks run on
multicore CPUs.

This paper includes the following contributions:
\begin{itemize}
\item A DSEL based on \Cxx to build parallel dataflow graphs for SDR signal
  processing, supporting loops, conditionals, pipeline and fork/join
  parallelism.
\item A set of micro-benchmarks to analyze the time taken by the different
  constructs;
\item A complete, real-life example using the DSEL, of a DVB-S2 transceiver
  running on both x86 and ARM CPUs.
\end{itemize}
Section~\ref{sec:related} discusses related works. The proposed DSEL is
presented in Section~\ref{sec:dsl}. Section~\ref{sec:scheduling} details
scheduling and parallelism supports. Section~\ref{sec:appli} experiments with a
DVB-S2 implementation built on the DSEL.

\section{Related Works}
\label{sec:related}

Many languages dedicated to streaming applications have been
introduced~\cite{Buck2004,Amarasinghe2005,Liao2006,Black-Schaffer2010,
Glitia2010,Thies2010,DeOliveiraCastro2017}. These languages are often variants
of the cyclo-static dataflow model and propose automatic parallelization
techniques such as pipelining and forks/joins.

In~\cite{Dardaillon2016}, authors proposed a full compilation chain for SDR,
based on LLVM on heterogeneous MPSoCs. This is promising but it differs from our
approach. Indeed, we chose to integrate our language into \Cxx, making the DSEL
compatible with any \Cxy{11} compilers. Works are also tackling OS and hardware
aspects of SDR~\cite{Tan2011,Li2014b,Hussain2018}. However, the studied SDR
systems are much simpler than those addressed in this paper.

Few solutions specifically target SDR sub-domain so far. GNU
Radio~\cite{GNURadio} is the most famous one. It is open source and largely
adopted by the community. It comes bundled with a large variety of digital
communication techniques used in real life systems.
The last version of GNU Radio (3.9) can take advantage of multi-core CPUs. One
thread is spawn per block and the scheduling is directly managed by the
operating system. While sufficient on uniform memory access (UMA)
architectures~\cite{Bloessl2019}, this does not take into account non uniform
memory access (NUMA) architectures with many cores. A drawback of assigning a
block per thread is that the designed SDR system is strongly linked to the
parallelism strategy. Depending on the CPU architecture, it can be necessary to
change the parallelism strategy while keeping the same SDR system description.
GNU Radio designers are currently working on a proof of concept scheduler
(newsched) for the future GNU Radio version 4~\cite{Muller2020}. They introduced
the concept of workers that can execute more than one block on a physical core.
To the best of our knowledge, this new version of GNU Radio breaks the
compatibility with the existing systems designed with GNU Radio and is not yet
fully implemented. However, this new version goes in the same direction as what
we propose and we hope that some contributions of this paper could help the GNU
Radio project. To the best of our knowledge, GNU Radio does not implement the
duplication mechanism presented in Section~\ref{sec:parallel} and we show this
is a key mechanism for high throughputs and scalability. Besides, a new
construct in the next section (cf. the switcher module in
Section~\ref{sec:compo}) allows the design loops and conditions for SDR systems,
not yet supported by GNU Radio where only static directed acyclic graph can be
managed.

Some other works are focusing particularly on an SDR implementation for a DVB-S2
transceiver. Hereafter are the projects we have identified:
\begin{itemize}
  \item \textbf{leansdr}. A standalone open source project~\cite{leansdr}. A
    low-density parity-check (LDPC) bit-flipping decoder~\cite{Ryan2009} is
    chosen. The project does not support multi-threading.
  \item \textbf{gr-dvbs2rx}. An open source out-of-tree module~\cite{gr-dvbs2rx}
    for GNU Radio. One of the motivation of this project is to increase the
    throughput compared to leansdr. The project is open-source and we were able
    to perform a fair comparison. The results are presented in
    Section~\ref{sec:comparison}.
  \item \textbf{Grayver and Utter}. In a recently published
    paper~\cite{Grayver2020}, they succeed in building a 10 Gb/s DVB-S2 receiver
    on a cluster of server-class CPUs. The implementation is closed sources,
    making fair comparisons difficult.
\end{itemize}

\section{Description of the Proposed Domain Specific Embedded Language}
\label{sec:dsl}

This section introduces a DSEL working on \emph{sets of symbols} (aka frames).
It implements a form of the dataflow model, single rate, tailored to the
relevant characteristics of digital communication chains with channel coding.
The language defines \emph{elementary} and \emph{parallel} components.

\subsection{Elementary Components}
\label{sec:compo}

Four elementary components are defined: \emph{sequence}, \emph{module},
\emph{task} and \emph{socket}. The \emph{task} is the fundamental component. It
can be an encoder, a decoder or a modulator for instance and is a
single-threaded code function. It is designated as \emph{filter} in the standard
dataflow model. Though unlike a dataflow filter, a task can have an internal
state and a private memory to store temporary data. Additionally, a set of tasks
can share a common internal/private memory. In that case, multiple tasks are
grouped into a single \emph{module}. The main problem with internal memory is
that tasks cannot be executed safely by several threads in parallel because of
data races. However, in many cases the expression of a task or a set of tasks
can be simplified by allowing stateful tasks and modules.

A task can consume and produce public data through the input and/or output
\emph{sockets} it exposes. Connecting the sockets of different tasks is called
\emph{binding}. An input socket can only be bound to one output socket, while an
output socket can be bound to multiple input sockets. A task can only be
executed once all its input sockets are bound.

Tasks can be grouped into a \emph{sequence}. A sequence corresponds to a static
schedule of tasks. To create a sequence, the designer specifies the first tasks
and the last tasks to execute. Then the connected tasks are analyzed and a
sequence object is built. The analysis is a depth-first traversal of the task
graph and independent tasks are ordered according to the binding order of their
inputs. The principle is to add a task to the array of function pointers when
all the input sockets are visited in the depth first traversal of the tasks
graph. After that, the output sockets of the current task are followed to reach
new tasks. The order in which the tasks have been traversed is memorized in the
sequence. When the designer calls the \texttt{exec} method on a sequence, the
tasks are executed successively according to this statically scheduled order.

\begin{figure}[htp]
  \centering
  \subfloat[][Simple chain sequence.\label{fig:sdr_dsl_sequence_chain}]{\includegraphics[scale=0.6]{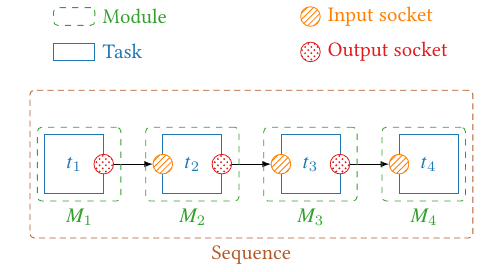}}
  \hfill
  \subfloat[][Sequence with multiple first and last tasks.\label{fig:sdr_dsl_sequence_generic}]{\includegraphics[scale=0.6]{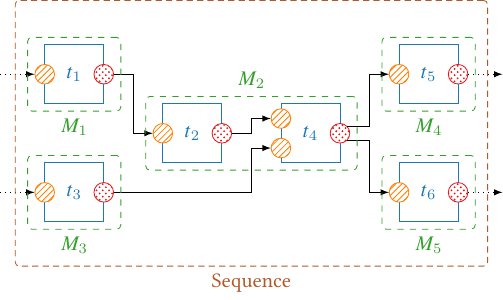}}
  \caption{Example of task sequences.}
  \label{fig:sdr_dsl_sequence}
\end{figure}

Fig.~\ref{fig:sdr_dsl_sequence} shows two examples of task sequences.
Fig.~\ref{fig:sdr_dsl_sequence_chain} is a simple chain of tasks. The designer
only needs to specify its first task ($t_1$); the sequence analysis then follows
the binding until the last task ($t_4$). In
Fig.~\ref{fig:sdr_dsl_sequence_generic}, bound tasks exist before and after the
current sequence, which also has two \emph{first} tasks ($t_1$ and $t_3$) and
two \emph{last} tasks ($t_5$ and $t_6$). In this case, the designer has to
explicitly specify that $t_1$ and $t_3$ are first tasks. If $t_1$ is
sequentially defined before $t_3$ then $t_1$ will be executed first and $t_3$
after. The analysis starts from $t_1$ and continue to traverse new tasks if
possible. In this example, $t_2$ can be executed directly after $t_1$, but $t_4$
cannot because it depends on $t_3$. So the analysis stops after $t_2$ and then
restarts from $t_3$. Actually, the index $i$ of the $t_i$ task represents the
execution order. The $t_5$ and $t_6$ last tasks have to be explicitly specified
because their output sockets are bound: the analysis cannot guess the end of the
sequence.

In targeted SDR applications, processing is continuously repeated on batches of
frames as long as the system is on. A sequence is thus executed in a loop. When
the last sequence task is executed, the next task is the first one on the next
frame. The designer can control whether the sequence should restart by supplying
a \emph{condition function} to the sequence \texttt{exec} method. The boolean
returned by the function conditions whether the sequence is repeated. A task of
a sequence may also raise an abort exception upon some condition, to immediately
stop the current sequence execution and start the first task of the sequence.
In Fig.~\ref{fig:sdr_dsl_sequence_chain} if the $t_3$ task
raises the abort exception then the next executed task is $t_1$.

Some digital communication systems include schemes that require a loop or a
conditional. A sequence of tasks is executed one or more times depending on a
\emph{condition} task (or a \emph{control} task). To build loops and
conditionals, we introduce a switcher module composed of two control flow tasks.
The \emph{select} task selects one among several exclusive input paths. The
\emph{commute} task creates two or more exclusive output paths.

\begin{figure*}[htp]
  \centering
  \subfloat[][Decomposition of the loop in modules.\label{fig:sdr_dsl_loop_mdl}]{\includegraphics[scale=0.6]{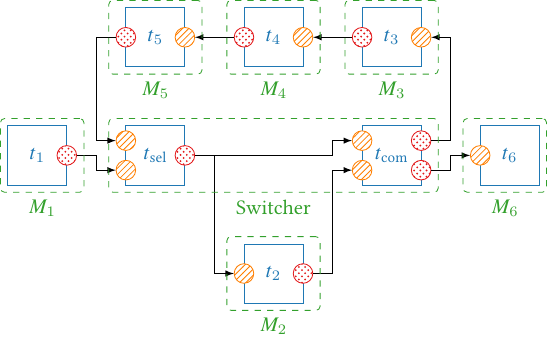}}
  \quad
  \subfloat[][Decomposition of the loop in sub-sequences.\label{fig:sdr_dsl_loop_ss}]{\includegraphics[scale=0.6]{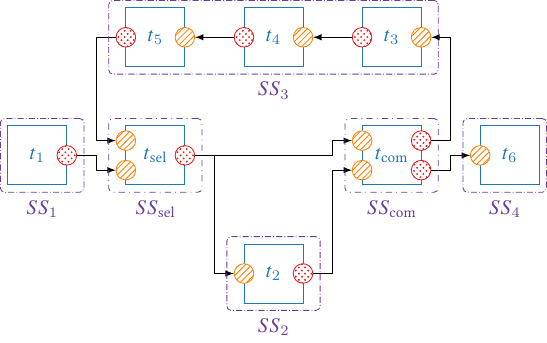}}
  \quad
  \subfloat[][Execution graph.\label{fig:sdr_dsl_loop_exe}]{\includegraphics[scale=0.6]{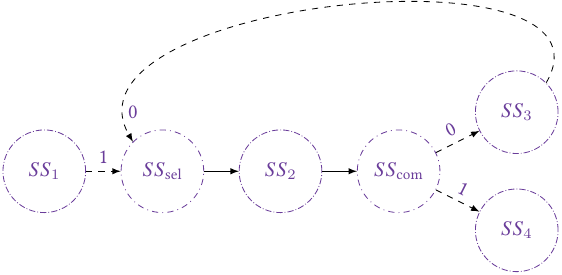}}
  \caption
    [Example of a sequence with a while loop.]
    {Example of a sequence with a while loop.}
  \label{fig:sdr_dsl_loop}
\end{figure*}

\begin{algorithm}[htp]
\caption{Pseudo code of a loop (corresponding to Fig.~\ref{fig:sdr_dsl_loop})}\label{alg:sdr_dsl_loop}
\begin{algorithmic}
\STATE execute $SS_1$;
\STATE \textbf{while} execute $SS_2$ \textbf{and not} $t_2$.out:
\STATE \hspace{0.5cm} execute $SS_3$;
\STATE execute $SS_4$;
\end{algorithmic}
\end{algorithm}

Fig.~\ref{fig:sdr_dsl_loop} illustrates a loop. To build a loop structure (a
while loop in the example), the \emph{select} and \emph{commute} tasks (in the
given order) of a common switcher module are used (see
Fig.~\ref{fig:sdr_dsl_loop_mdl}). By convention, in a switcher module, the
selected path is initialized to the highest possible path (here 1). So, at the
first $t_\text{sel}$ execution, the $t_1$ output will be selected. Then the
$t_2$ control task will send 0 or 1 as a control socket to $t_\text{com}$.
Here the $t_2$ loop control task is based on the $t_\text{sel}$ output socket.
As a consequence, the path selection (0 or 1) is dynamic and depends on the
runtime data. It is also possible to model the \texttt{for} loop behavior by
ignoring the $t_2$ input data and by adding an internal state to $M_2$, namely
the loop counter. If $t_\text{com}$ receives a 0, then the internal path of the
switcher will be 0. Then $t_3$, $t_4$ and $t_5$ tasks will be executed and
$t_\text{sel}$ will select the $t_5$ output instead of the $t_1$ output, and so
on. Fig.~\ref{fig:sdr_dsl_loop_ss} shows how the tasks are regrouped into
sub-sequences. It enables to build the execution graph illustrated in
Fig.~\ref{fig:sdr_dsl_loop_exe}. The corresponding pseudo code of the presented
loop is shown in Alg.~\ref{alg:sdr_dsl_loop}. One can note that in the
proposed DSEL there is no limitation to include nested loops schemes. The loop
pattern is common in iterative demodulation/decoding. This is why it is required
in a DSEL dedicated to SDR. As an exception rule in the graph construction, the
\emph{select} task is added to the graph when its last input socket is visited
(while all the other tasks require all input sockets to be visited).

\begin{figure}[htp]
  \centering
  \subfloat[][Decomposition of the switch in sub-sequences.\label{fig:sdr_dsl_swi_ss}]{\includegraphics[scale=0.6]{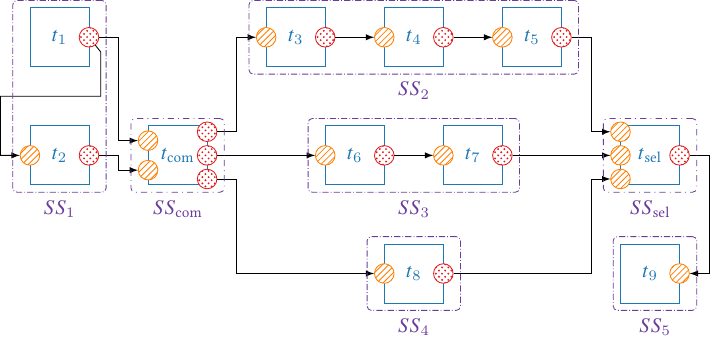}}
  \hfill
  \subfloat[][Execution graph.\label{fig:sdr_dsl_swi_exe}]{\includegraphics[scale=0.6]{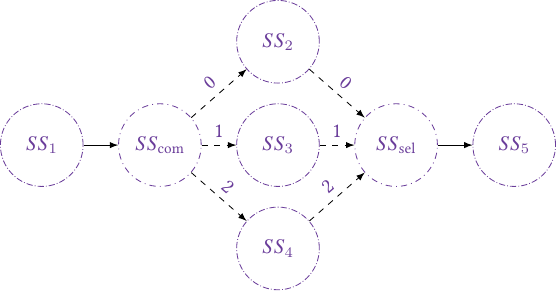}}
  \caption
    [Example of a sequence with a switch.]
    {Example of a sequence with a switch.}
  \label{fig:sdr_dsl_swi}
\end{figure}

\begin{algorithm}[htp]
\caption{Pseudo code of a switch (corresponding to Fig.~\ref{fig:sdr_dsl_swi})}\label{alg:sdr_dsl_swi}
\begin{algorithmic}
\STATE execute $SS_1$;
\STATE \textbf{switch} $t_2$.out:
\STATE \hspace{0.5cm} \textbf{case} 0: execute $SS_2$;
\STATE \hspace{0.5cm} \textbf{case} 1: execute $SS_3$;
\STATE \hspace{0.5cm} \textbf{case} 2: execute $SS_4$;
\STATE execute $SS_5$;
\end{algorithmic}
\label{alg1}
\end{algorithm}

Fig.~\ref{fig:sdr_dsl_swi} illustrates a \texttt{switch} structure. The same
switcher module presented in while loop structures is necessary. The number of
output/input sockets in resp. $t_\text{com}$/$t_\text{sel}$ tasks is 3 instead
of 2 in the while loop example. Also, the position of these tasks has been
swapped, in the current example $t_\text{com}$ is executed before
$t_\text{sel}$. $t_2$ is a control task that depends on the output of $t_1$. The
$t_2$ task output can be 0, 1 or 2. The switch exclusive path is determined
dynamically depending on the runtime data. Fig.~\ref{fig:sdr_dsl_swi_ss} shows
the decomposition of the tasks in sub-sequences and
Fig.~\ref{fig:sdr_dsl_swi_exe} presents the resulting execution graph.
Alg.~\ref{alg:sdr_dsl_swi} gives the corresponding pseudo code. The switch
pattern is useful in many SDR contexts. For instance, depending on the signal to
noise ratio (SNR), the receiver can select a different path adapted to the
signal quality.

\subsection{Performance Evaluation on Micro-benchmarks}

In this section, an estimation of the DSEL overhead is measured from four
micro-benchmarks: a simple chain (see Fig.~\ref{fig:sdr_dsl_sequence_chain})
denoted $MB_1$, a single \texttt{for} loop (Fig.~\ref{fig:sdr_dsl_loop}) denoted
$MB_2$, a system of two nested \texttt{for} loops denoted $MB_3$, and a system
with a \texttt{switch} (Fig.~\ref{fig:sdr_dsl_swi}) denoted $MB_4$. In $MB_1$,
$MB_2$ and $MB_3$ three computational tasks are chained. In $MB_2$, the loop
performs 10 iterations. In $MB_3$, the inner loop performs 5 iterations, the
outer loop performs 2 iterations. In $MB_4$, three computational tasks are
chained in the first path, two in the second path and a single in the last
path. Moreover, an iterate task is configured to perform a cyclic path selection
(0,1,2,0,1,2,...). In each computational task, an active wait of the same amount
of time is performed. Four types of tasks are used: \emph{computational} tasks
$\mathcal{C}$, \emph{select} and \emph{commute} switcher module tasks
$\mathcal{S}_\text{sel}$ and $\mathcal{S}_\text{com}$ resp., and \emph{iterate}
tasks $\mathcal{I}$ to determine paths in loops and switches ($\mathcal{I}$ =
control task).

Evaluations ran on a single core of an Intel\R Core\TM i5-8250U @ 1.60 GHz. The
\emph{Turbo Boost} mode has been disabled. This processor has a 15-Watt TDP that
matches embedded system constraints. Though duration of a $\mathcal{C}$ task is
controlled by the programmer, we measured a constant 135~ns overhead due to the
DSEL and to the system call behind the
\texttt{std::chrono::steady\_clock::now()} function. We measured
$\mathcal{S}_\text{sel}$ tasks around 60~ns, $\mathcal{S}_\text{com}$ tasks
around 80~ns, and $\mathcal{I}$ tasks around 70~ns. Later on,
$\mathcal{S}_\text{sel}$, $\mathcal{S}_\text{com}$ and $\mathcal{I}$ tasks are
reported as overhead. both $\mathcal{S}_\text{sel}$ and $\mathcal{S}_\text{com}$
tasks are copy-less, thus for a given configuration, their execution time is
constant.

\begin{table*}[htp]
  \centering
  \caption
    {Execution of 1 125 000 $\mathcal{C}$ tasks of 4 $\mu s$ each. Theoretical
     execution time is 4500 ms for each micro-benchmark.}
  \label{tab:sdr_latency}
  \begin{footnotesize}
  \begin{tabular}{c r r r r r r r r r r r}
    \toprule
            &            &          & \multicolumn{9}{c}{\text{Overhead}} \\ \cmidrule(lr){4-12}
            &            &          & \multicolumn{2}{c}{$\mathcal{C}$ tasks} & \multicolumn{2}{c}{$\mathcal{S}_\text{sel}$ tasks} & \multicolumn{2}{c}{$\mathcal{S}_\text{com}$ tasks} & \multicolumn{2}{c}{$\mathcal{I}$ tasks} & \multicolumn{1}{c}{Other} \\
    \cmidrule(lr){4-5}  \cmidrule(lr){6-7} \cmidrule(lr){8-9} \cmidrule(lr){10-11} \cmidrule(lr){12-12}
    Label   & Seq. exec. & Run time (ms) &   Exec. &   Time (ms) &  Exec. &  Time (ms) &  Exec. &  Time (ms) &  Exec. &  Time (ms) &  Time (ms) \\
    \midrule
    $MB_1$  & 375000     &  4656.45 & 1125000 & 151.86 &     -- &    -- &    --  &    -- &     -- &    -- &  4.59 \\
    $MB_2$  &  37500     &  4744.08 & 1125000 & 151.86 & 412500 & 24.75 & 412500 & 33.00 & 412500 & 28.88 &  5.59 \\
    $MB_3$  &  37500     &  4777.03 & 1125000 & 151.86 & 562500 & 33.75 & 562500 & 45.00 & 562500 & 39.38 &  7.04 \\
    $MB_4$  & 562500     &  4784.88 & 1125000 & 151.86 & 562500 & 33.75 & 562500 & 45.00 & 562500 & 39.38 & 14.89 \\
    \bottomrule
  \end{tabular}
  \end{footnotesize}
\end{table*}

Tab.~\ref{tab:sdr_latency} reports the execution time of 1~125~000~$\mathcal{C}$
tasks for each case. Column \emph{Seq. exec.} gives the number of sequence
executions required to run 1~125~000~$\mathcal{C}$ tasks. Theoretical time is
computed directly from the number of $\mathcal{C}$ tasks and the duration of the
active waiting in each task: $\mathcal{T}_\text{theoretical} = 1125000 \times
4~\mu\text{s} = 4500~\text{ms}$.  \emph{Run time} column reports the measured
execution time. Remaining columns report task counts and overheads per task
types. Last column \emph{Other} reports the residual time that does not come
from the tasks execution. In each benchmark, a stop condition is evaluated at
the end of the sequence. The condition checks that the current number of
executions is lower than the one given in the  \emph{Seq. exec.} column. This
comes with an extra cost because of an additional function call for each
sequence execution. This is also why the execution time of $MB_4$ is higher than
$MB_3$, there is significantly more sequence executions in $MB_4$.

\begin{figure}[htp]
  \centering
  \includegraphics[width=0.8\linewidth]{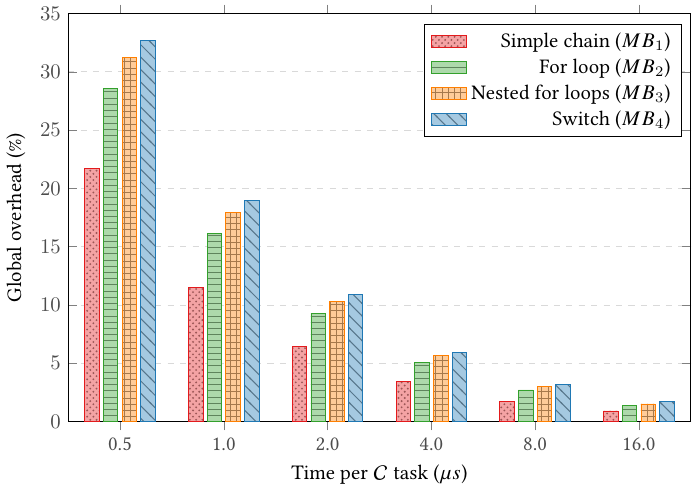}
  \caption
    [Global overhead of the proposed DSEL on 4 micro-benchmarks.]
    {Global overhead of the proposed DSEL on 4 micro-benchmarks.}
  \label{fig:sdr_dsl_overhead}
\end{figure}

Fig.~\ref{fig:sdr_dsl_overhead} shows the overhead depending on the granularity
of the $\mathcal{C}$ tasks. It results that from 4 $\mu$s tasks, the proposed
DSEL has an acceptable overhead. For tasks longer than 4 $\mu$s the overhead is
negligible. This shows that the proposed DSEL matches the low latency
requirements of SDR systems.

\subsection{Parallel Components}
\label{sec:parallel}

\begin{figure}[htp]
  \centering
  \includegraphics[scale=0.6]{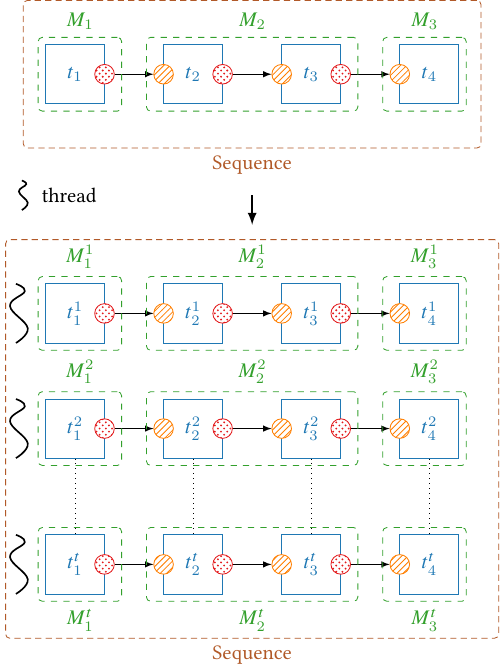}
  \caption
    [Sequence duplication for multi-threaded execution.]
    {Sequence duplication for multi-threaded execution. The modules are
     duplicated with their tasks and data.}
  \label{fig:sdr_dsl_sequence_dup}
\end{figure}

A sequence can be duplicated, to let several threads execute it in
parallel, see Fig.~\ref{fig:sdr_dsl_sequence_dup}. The number $t$ of duplicates
is a parameter of its constructor. There is no synchronization between sequence
duplicates. Each threaded sequence can be executed on one dedicated core and the
public data transfers remain on this core for the data reuse in the caches. By
default, modules have no duplication mechanism (see next
section).

In some particular cases such as in the signal synchronization processing, the
tasks can have a dependency on themselves. It is then impossible to duplicate
the sequence because of the sequential nature of the tasks. To overcome this
issue, the well-known pipelining strategy can be applied to increase the
sequence throughput up to the slowest task throughput. The proposed DSEL comes
with a specific \emph{pipeline} component to this purpose. The pipeline takes
multiple sequences as input. Each sequence of the pipeline is called a
\emph{stage}, run on one thread. For instance, a 4-stage pipeline creates
4~threads. A pipeline stage can be combined with the sequence duplication
strategy. It means that there are nested threads in the current stage thread.
Pipelining comes with an extra synchronization cost between the stage threads,
implementation details are discussed in the next section.

\section{Automated Parallelization Techniques}
\label{sec:scheduling}

\subsection{Sequence Duplication}

In a fully dataflow-compliant model, there is no need to duplicate the sequence
because a stateless task is always thread-safe. In the proposed DSEL with
stateful tasks, a \emph{clone} method is defined for each module to deal with
internal state and private memory (stored in the module). The \emph{clone}
method is polymorphic and defined in the \emph{Module} abstract class. It relies
on the implicit copy constructors and a \emph{deep copy} protected method
(overridable). It is the responsibility of the \emph{ModuleImpl} developer to
correctly override the \emph{deep copy} method and to make sure the duplication
is valid for this module. The \emph{deep copy} method deals with pointer and
reference members. If the pointer/reference members are read-only
(\texttt{const}), then the implicit copy constructor copies the memory addresses
automatically. When the current \emph{ModuleImpl} class owns one ore more
writable references, the module cannot be cloned and its tasks are sequential.
However, for a writable pointer member, the developer can explicitly allocate a
new pointer in the \emph{deep copy} method.

\subsection{Pipeline}

\begin{figure}[htp]
  \centering
  \subfloat[][Description of a pipeline: tasks creation, tasks binding and
              sequences/stages definition (with the corresponding number of
              threads).\label{fig:sdr_dsl_pipeline_usr}]{
    \includegraphics[scale=0.58]{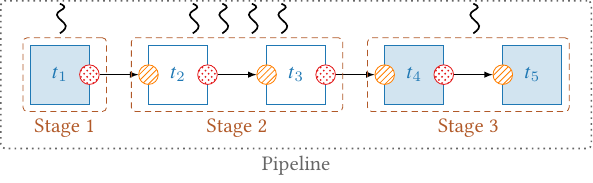}}
  \\
  \subfloat[][Automatic parallelization of a pipeline description: sequence
              duplications, 1 to $n$ and $n$ to 1 adaptors creation and binding.\label{fig:sdr_dsl_pipeline_adp}]{
    \includegraphics[scale=0.58]{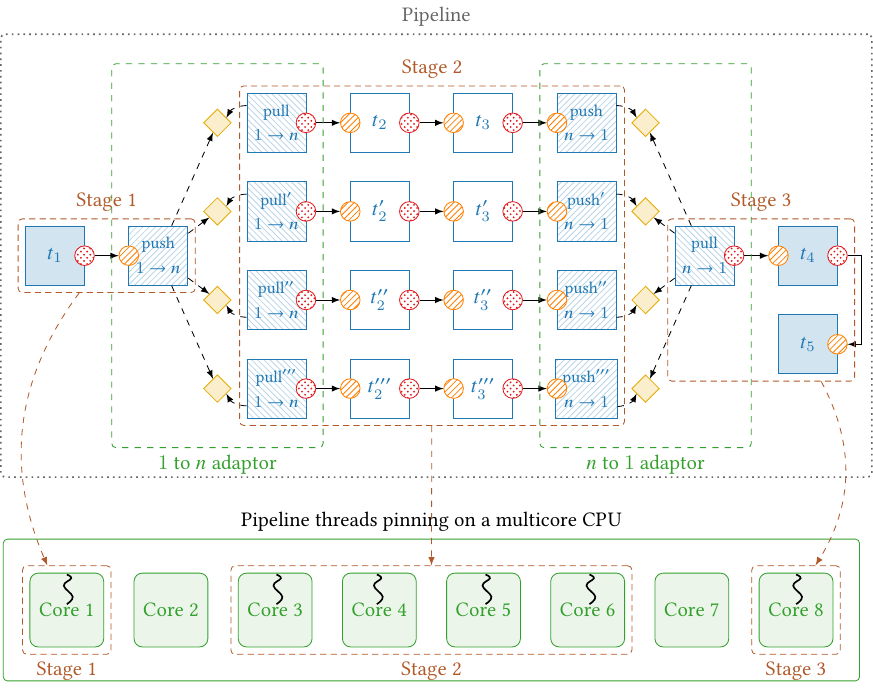}}
  \caption{Example of a pipeline description and the associate transformation
    with adaptors.}
  \label{fig:sdr_dsl_pipeline}
\end{figure}

\begin{figure}[htp]
  \begin{lstlisting} // 1) creation of the module objects
 module::M1 m1_obj(/* ... */); // 'M1' class & 't1' task
 module::M2 m2_obj(/* ... */); // 'M2' class & 't2' task
        . . .
 module::M5 m5_obj(/* ... */); // 'M5' class & 't5' task

 // 2) binding of the tasks
 m2_obj[sck::t2::in] = m1_obj[sck::t1::out];
 m3_obj[sck::t3::in] = m2_obj[sck::t2::out];
        . . .
 m5_obj[sck::t5::in] = m4_obj[sck::t4::out];

 // 3) creation of the pipeline (= sequences and pipeline
 //    analyses)
 tools::Pipeline pipeline(
   // first task of the sequence (for validation purpose)
   m1_obj[tsk::t1],
   // description of the sequence decomposition in
   // 3 pipeline stages
   {
     { { m1_obj[tsk::t1] },   // first tasks of stage 1
       { m1_obj[tsk::t1] } }, // last  tasks of stage 1
     { { m2_obj[tsk::t2] },   // first tasks of stage 2
       { m3_obj[tsk::t3] } }, // last  tasks of stage 2
     { { m4_obj[tsk::t4] },   // first tasks of stage 3
       { m5_obj[tsk::t5] } }, // last  tasks of stage 3
   },
   // number of threads per stage (4 sequence duplications
   // in stage 2)
   { 1,4,1 }, /* ... */
   // explicit pinning of the threads
   {
     {       1 }, // stage 1: thread  '1'   to core  '1'
     { 3,4,5,6 }, // stage 2: threads '1-4' to cores '3-6'
     {       8 }, // stage 3: thread  '1'   to core  '8'
 });

 // 4) execution of the pipeline, it is indefinitely
 //    executed in loop
 pipeline.exec([]() { return false; });\end{lstlisting}
  \caption{\Cxx DSEL source code of the pipeline described in
    Fig.~\ref{fig:sdr_dsl_pipeline}.}
  \label{lst:sdr_dsl_pipeline}
\end{figure}

In this section, the pipeline implementation is illustrated through a simple
example. Fig.~\ref{fig:sdr_dsl_pipeline} shows the difference between a
pipeline description (see Fig.~\ref{fig:sdr_dsl_pipeline_usr}) and its actual
instantiation (see Fig.~\ref{fig:sdr_dsl_pipeline_adp}). In
Fig.~\ref{fig:sdr_dsl_pipeline} we suppose that the $t_1$, $t_4$ and $t_5$ tasks
cannot be duplicated (plain boxes). The designer knows that the execution
time of the $t_1$ task is higher than the cumulated execution time of tasks
$t_4$ and $t_5$. We assume that the cumulated execution time of $t_2$ and $t_3$
is approximatively four times higher than $t_1$. This knowledge motivates the
splitting of the stages 1, 2 and 3. There is no need to split the $t_4$ and
$t_5$ tasks in two stages because the overall throughput is limited by the
slowest stage ($t_1$ here). Stage 2 is duplicated four times to increase its
throughput by four as we know that its latency is approximatively four times
that of Stage 1. In general, a preliminary profiling phase of the sequential
code is required to guide the pipeline strategy.
Listing~\ref{lst:sdr_dsl_pipeline} presents the \Cxx DSEL source code
corresponding to the pipeline description in
Fig.~\ref{fig:sdr_dsl_pipeline_usr}. Each task $t_i$ is contained (as a method)
in the $M_i$ module (or class). The four main steps are: 1) Creation of the
modules; 2) Binding of the tasks; 3) Creation of the pipeline strategy;
4) Pipeline execution.

Fig.~\ref{fig:sdr_dsl_pipeline_adp} presents the internal structure of the
pipeline. As we can see, new tasks have been automatically added:
$push_{1 \rightarrow n}$, $pull_{1 \rightarrow n}$ shared by a \emph{1 to n
adaptor} module and $push_{n \rightarrow 1}$, $pull_{n \rightarrow 1}$ shared by
a \emph{n to 1 adaptor} module. The binding as been modified to insert the tasks
of the adaptors. In the initial pipeline description, $t_1$ is bound to $t_2$.
In a parallel pipelined execution this is not possible anymore because many
threads are running concurrently: One for stage 1, four for stage 2 and
one for stage 3 in the example. To this purpose, the adaptors implement a
producer-consumer scheme. The yellow diamonds represent the buffers that are
required. The $push_{1 \rightarrow n}$ and $pull_{n \rightarrow 1}$ tasks can
only be executed by a single thread while the $pull_{1 \rightarrow n}$ and
$push_{n \rightarrow 1}$ tasks are thread-safe. The $push_{1 \rightarrow n}$
task copies its input socket in one buffer each time it is called. There is one
buffer per duplicated sequence. To guarantee that the order of the
input frames is preserved, a round-robin scheduling has been adopted. On the
other side, the $pull_{n \rightarrow 1}$ task is copying the data from the
buffers to its output socket, with the same round-robin scheduling.

The size of the synchronization buffers in the adaptors are defined on creation.
The default size is one. During the copy of the input socket data in one of the
buffers, threads cannot access the data until the copy is finished. The
synchronization is automatically managed by the framework. If the buffer is
full, the producer ($push_{1 \rightarrow n}$ and $push_{n \rightarrow 1}$ tasks)
has to wait. The same applies for the consumer ($pull_{1 \rightarrow n}$ and
$pull_{n \rightarrow 1}$ tasks) if the buffer is empty. We implemented both
active and passive waiting.

Copies from and to buffers are expensive. These copies are removed by
dynamically re-binding the tasks just before and just after the \emph{push} and
\emph{pull} tasks, and casting the tasks into copyless variants. It is also
necessary to bypass the regular execution in the $push_{1 \rightarrow n}$,
$pull_{1 \rightarrow n}$, $push_{n \rightarrow 1}$ and $pull_{n \rightarrow 1}$
tasks. This replaces the source code of the data buffer copy by a simple pointer
copy. The pointers are exchanged cyclically.

In Fig.~\ref{fig:sdr_dsl_pipeline_adp}, the pipeline threads are pinned to
specific CPU cores. This is the direct consequence of the lines 31-35 in
Listing~\ref{lst:sdr_dsl_pipeline}. The \emph{hwloc}
library~\cite{Broquedis2010} has been used and integrated in our DSEL to pin the
software threads to processing units (hardware threads). In the given
example, we assume that the CPU cores can only execute one hardware thread (SMT
off). The threads pinning is given by the designer. This can improve the
multi-threading performance on NUMA architectures.

Fig.~\ref{fig:sdr_dsl_pipeline} is an example of a simple chain of tasks. More
complicated task graphs can have more than two tasks to synchronize between two
pipeline stages. The adaptor implementation can manage multiple socket
synchronizations. The key idea is to deal with a 2-dimensional array of buffers.
Another difficult case is when a task $t_1$ is in stage 1 and possesses an
output socket bound to an other task $t_{x}$ which is located in the stage 4.
To work, the pipeline adaptors between the stages 1 and 2 and the stages 2 and 3
automatically synchronize the data of the $t_1$ output socket.

\section{Application on the DVB-S2 Standard}
\label{sec:appli}

In this section, we present a real use case of our DSEL. The second generation
of Digital Video Broadcasting standard for Satellite (DVB-S2)~\cite{ETSI2005} is
a flexible standard designed for broadcast applications. DVB-S2 is typically
used for the digital television (HDTV with H.264 source coding). The full DVB-S2
transmitter and receiver are implemented in a SDR-compliant system. Two
Universal Software Radio Peripherals (USRPs) N320\footnote{USRP N320:
\url{https://www.ettus.com/all-products/usrp-n320/}.} have been used for the
analog signal transmission and reception where all the digital processing of the
system have been implemented. The purpose of this section is not to detail all
the implemented tasks extensively, but rather to expose the system as a whole.
Some specific focuses are given to describe the main encountered problems and
solutions.

\subsection{Transmitter Software Implementation}

\begin{table}[htp]
  \centering
  \caption
    [Selected DVB-S2 configurations (MODCOD).]
    {Selected DVB-S2 configurations (MODCOD).}
  \label{tab:sdr_dvbs2_modcod}
  {\resizebox{\linewidth}{!}{
  \begin{tabular}{c c c c c c c}
    \toprule
    Config.  & Mod. & Rate $R$ & $K_\text{BCH}$ & $K_\text{LDPC}$ & Interleaver\\
    \midrule
    MODCOD 1 &  QPSK & 3/5 &  9552 &  9720 & no\\
    MODCOD 2 &  QPSK & 8/9 & 14232 & 14400 & no\\
    MODCOD 3 & 8-PSK & 8/9 & 14232 & 14400 & col/row\\
    \bottomrule
  \end{tabular}
  }}
\end{table}

The DVB-S2 coding scheme rests upon the serial concatenation of a Bose,
Ray-Chaudhuri \& Hocquenghem (BCH) and a LDPC code. The selected modulation is a
Phase-Shift Keying (PSK). The standard defines 32 MODulation and CODing schemes
or \emph{MODCODs}. This work focuses on the 3~MODCODs given in
Tab.~\ref{tab:sdr_dvbs2_modcod}. Depending on the MODCOD, the PSK modulation and
the LDPC code rate $R$ vary. In MODCOD~1 and 2 there is no interleaver, MODCOD 3
uses a column/row interleaver. $K_\text{BCH}$ or $K$ is the number of
information bits and the input size of the BCH encoder. $N_\text{BCH}$ or
$K_\text{LDPC}$ is the output size of the BCH encoder and the input size of the
LDPC encoder. For each selected MODCOD, $N_\text{LDPC} = 16200$. With the pay
load header (PLH) and pilots bits, the frame size ($N_\text{PLH}$ or $N$)
contains a total of 16740 bits.

\begin{figure*}[htp]
  \centering
  \includegraphics[scale=0.45]{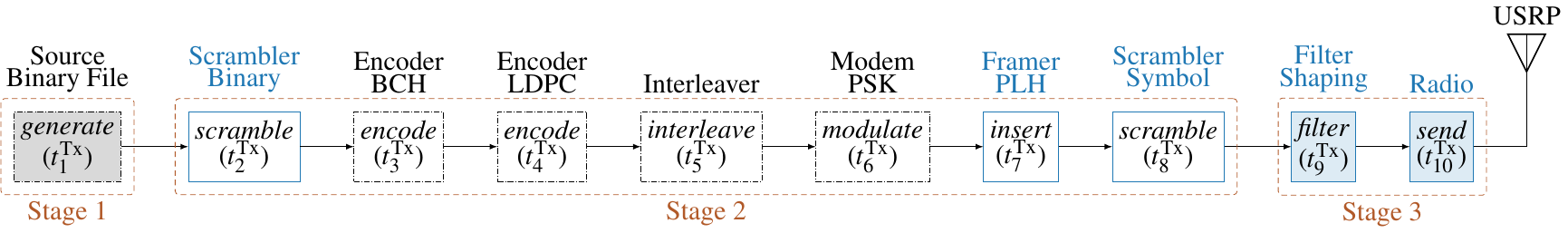}
  \caption
    [DVB-S2 transmitter software implementation.]
    {DVB-S2 transmitter software implementation.}
  \label{fig:sdr_dvbs2_transmitter}
\end{figure*}

Fig.~\ref{fig:sdr_dvbs2_transmitter} shows the DVB-S2 transmitter decomposition
in tasks and pipeline stages. Intrinsically sequential tasks are represented by
plain boxes. The DVB-S2 transmitter has been implemented in software with the
proposed DSEL. Out of conciseness, it is not detailed in this paper as it is
much more simpler than the receiver part of the system in terms of computational
requirement and complexity of the tasks graph.

\subsection{Receiver Software Implementation}

\begin{figure*}[htp]
  \centering
  \subfloat[][Waiting phase and learning phase 1 \& 2.\label{fig:sdr_dvbs2_receiver_learning}]{
    \includegraphics[scale=0.45]{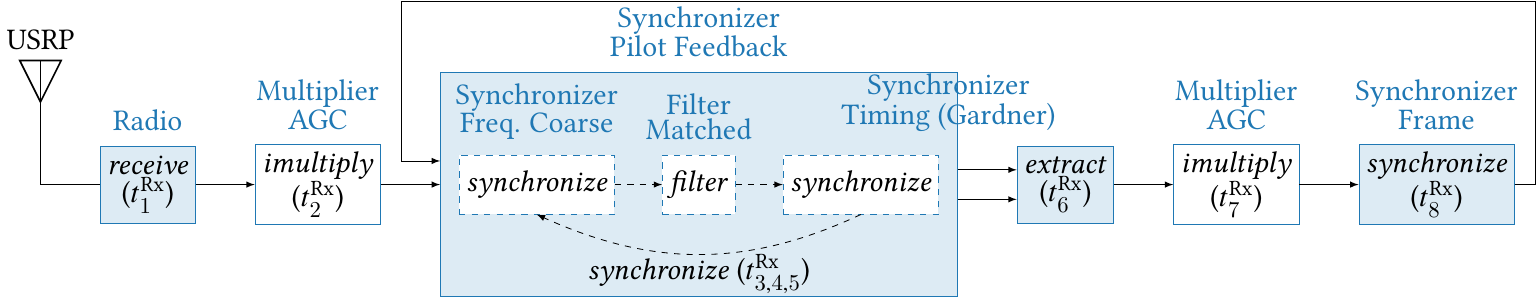}}
  \\
  \subfloat[][Learning phase 3 \& transmission phase.\label{fig:sdr_dvbs2_receiver_transmission}]{
    \includegraphics[scale=0.45]{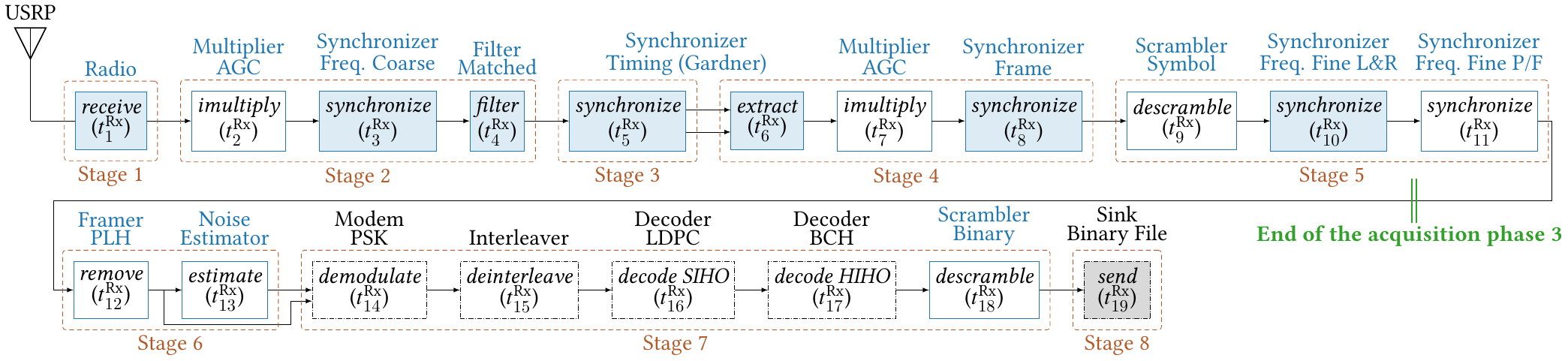}}
  \caption
    [DVB-S2 receiver software implementation.]
    {DVB-S2 receiver software implementation.}
  \label{fig:sdr_dvbs2_receiver}
\end{figure*}

Fig.~\ref{fig:sdr_dvbs2_receiver} presents the task decomposition of the
DVB-S2 receiver software implementation with the five distinct phases. The plain
tasks are intrinsically sequential and cannot be duplicated. The first
phase is called the \emph{waiting phase} (see
Fig.~\ref{fig:sdr_dvbs2_receiver_learning}). It consists in waiting until a
transmitter starts to transmit. The \emph{Synchronizer Frame} task
($t^\text{Rx}_8$) possesses a frame detection criterion. When a signal is
detected, the \emph{learning phase 1} (see
Fig.~\ref{fig:sdr_dvbs2_receiver_learning}) is executed during 150 frames.
After that the \emph{learning phase 2} (see
Fig.~\ref{fig:sdr_dvbs2_receiver_learning}) is also executed during 150
frames. After the \emph{learning phase 1 and 2}, the tasks have to be re-bound
for the \emph{learning phase 3} (see
Fig.~\ref{fig:sdr_dvbs2_receiver_transmission}). This last learning phase is
applied over 200 frames. After the 500 frames of these successive learning
phases, the final \emph{transmission phase} is established (see
Fig.~\ref{fig:sdr_dvbs2_receiver_transmission}).

In a real life communication systems, the internal clocks of the radios can
drift slightly. A specific processing has to be added in order to be resilient.
This is achieved by the \emph{Synchronizer Timing} tasks ($t^\text{Rx}_5$ and
$t^\text{Rx}_6$). Similarly, the radio transmitter frequency does not perfectly
match the receiver frequency, so the \emph{Synchronizer Frequency} tasks
($t^\text{Rx}_3$, $t^\text{Rx}_{10}$ and $t^\text{Rx}_{11}$) recalibrate the
signal to recover the transmitted symbols. Finally LDPC decoder is a block
coding scheme that requires to know precisely the first and last bits of the
codeword. The \emph{Synchronizer Frame} task ($t^\text{Rx}_8$) uses the PLH and
pilots bits inserted by the transmitter to recover the first and last symbols.
The \emph{Synchronizer Timing} module is composed by two separated tasks
(\emph{synchronize} or $t^\text{Rx}_5$ and \emph{extract} or $t^\text{Rx}_6$).
This behavior is different from the other \emph{Synchronizer} modules. The
\emph{synchronize} task ($t^\text{Rx}_5$ or $t^\text{Rx}_{3,4,5}$) has two
output sockets, one for the regular data and another one for a mask. The regular
data and the mask are then used by the \emph{extract} task ($t^\text{Rx}_5$) to
screen which data is selected for the next task. The \emph{Synchronizer Timing}
tasks ($t^\text{Rx}_5$ and $t^\text{Rx}_6$) have a high latency compared to the
others tasks, thus splitting the treatment in two tasks is a way to increase the
throughput of the pipeline.

Besides in some cases the task does not have enough samples to produce a frame.
In such cases, the \emph{extract} task raises an abort exception. The exception
is caught and the sequence restarts from the first task ($t^\text{Rx}_1$).

During the waiting and learning phases 1 and 2, the \emph{Synchronizer Freq.
Coarse}, the \emph{Filter Matched} and a part of the \emph{Synchronizer Timing}
have to work symbol by symbol. They have been grouped in the \emph{Synchronizer
Pilot Feedback} task ($t^\text{Rx}_{3,4,5}$). $t^\text{Rx}_{3,4,5}$ also
requires a feedback input from the \emph{Synchronizer Frame} task
($t^\text{Rx}_8$). This behavior is no longer necessary in subsequent phases, so
the $t^\text{Rx}_{3,4,5}$ task has been split in $t^\text{Rx}_3$,
$t^\text{Rx}_4$ and $t^\text{Rx}_5$. Consequently, the feedback from the
$t^\text{Rx}_8$ second output socket is left unbound.

\begin{figure}[htp]
  \centering
  \includegraphics[width=0.9\linewidth]{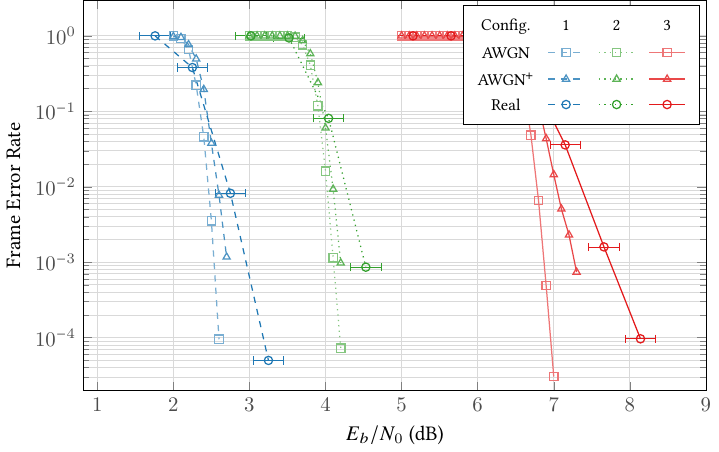}
  \caption
    [DVB-S2 frame error rate (FER) decoding performance.]
    {DVB-S2 frame error rate (FER) decoding performance (LDPC BP h-layered, min-sum, 10 ite.).}
  \label{fig:sdr_dvbs2_bfer}
\end{figure}

Fig.~\ref{fig:sdr_dvbs2_bfer} shows the frame error rate (FER) decoding
performance results of the 3 selected MODCODs. The shapes represent the channel
conditions: Squares stand for a standard simulated additive white Gaussian noise
(AWGN) channel, triangles are a simulated AWGN channel in which frequency shift,
phase shift and symbol delay have been taken into account, circles are the real
conditions measured performances with the USRPs. There is a  0.2 dB inaccuracy
in the noise estimated by the $t^\text{Rx}_{13}$ task. It is symbolized by the
extra horizontal bars over the circles. The MODCOD~1 is represented by dashed
lines, MODCOD~2 by dotted lines and MODCOD~3 by solid lines. For each MODCOD,
the LDPC decoder is based on the belief propagation algorithm with horizontal
layered scheduling (10 iterations) and with the min-sum node update rules. Each
DVB-S2 configuration has a well-separated SNR predilection zone.

\subsection{Open Source Integration with \AFFECT Toolbox}

The proposed software implementation of the DVB-S2 digital transceiver is open
source\footnote{DVB-S2 digital transceiver repository:
\url{https://github.com/aff3ct/dvbs2}.}.
It is described with the help of the \AFFECT toolbox~\cite{Cassagne2019a}.
\AFFECT is a library dedicated to the digital communication systems and more
specifically to the channel decoding algorithms. In this paper, we extend
\AFFECT with the presented DSEL to the SDR use case while keeping the
interoperability, reproducibility and maintainability philosophy initiated in
the toolbox. Some components are directly used from the \AFFECT library (black
dashed-dotted tasks in Fig.~\ref{fig:sdr_dvbs2_transmitter} and
Fig.~\ref{fig:sdr_dvbs2_receiver_transmission}) and are optimized for
efficiency. For instance, knowing that the LDPC decoding is one of the most
compute intensive task, an existing high performance SIMD implementation is
used, based on the portable \MIPP library~\cite{Cassagne2018}. Additional
\AFFECT tasks have been implemented specifically for this project (blue
boxes in Fig.~\ref{fig:sdr_dvbs2_transmitter},
~\ref{fig:sdr_dvbs2_receiver_learning}
and~\ref{fig:sdr_dvbs2_receiver_transmission}). These new tasks  mainly
address two areas: signal synchronizations and filters, and real-time
communications.

\subsection{Evaluation}

This section evaluates the receiver part of the system. The transmitter part as
it is not the most compute intensive part and high throughputs are much more
easier to reach. All the presented results have been obtained on two high-end
NUMA machines. One is composed by two Intel\R Xeon\TM Platinum 2.70 Ghz 8168
CPUs, 24 cores 128GB RAM (denoted as x86). \emph{Turbo Boost} mode has been
disabled for the reproducibility of the experiment results. Each core is powered
by AVX-512F SIMD ISA. The second architecture is composed by two Cavium
ThunderX2\R 2.00 GHz CN9975 v2.1 CPUs, 28 cores, 256 GB of RAM (denoted as ARM).
Each core is powered by NEON SIMD ISA. In the proposed implementation, the data
are represented by 32-bit floating-point numbers. Data parallelism level is thus
16 for AVX-512F ISA and 4 for NEON ISA. For both targets, the GNU C++ compiler
version 9.3 has been used with the following flags: \verb|-O3 -march=native|.

An high performance LDPC decoder implementation with the inter-frame SIMD
technique is used (the early termination criterion has been switched on). This
choice has the effect of computing sixteen/four frames at once in each task of
the receiver (depending on the x86 or ARM target). It negatively affects the
overall latency of the system (by a factor of sixteen/four). But it is not
important in the video streaming targeted application. The \emph{Decoder LDPC}
task ($t^\text{Rx}_{16}$) is the only one in the receiver to take advantage of
the inter-frame SIMD technique. The other tasks simply process sixteen/four
frames sequentially.

\begin{table}[t]
  \centering
  \caption
    {Tasks sequential throughputs and latencies of the DVB-S2 receiver
     (transmission phase, 16288 frames, inter-frame level = 16, MODCOD 2,
     error-free SNR zone, x86 target)
     Sequential tasks are represented by \colorbox{Paired-1!15}{blue} rows.
     The slowest seq. stage is in \colorbox{Paired-5!15}{red} while the
     slowest of all is in \colorbox{Paired-7!15}{orange}.}
  \label{tab:sdr_dvbs2_tasks_thr_lat}
  {\resizebox{\linewidth}{!}{
  \begin{tabular}{r r r r r}
    \toprule
                                                                             Stages and Tasks &  Throughput &   Latency & Time  \\
                                                                                              &      (Mb/s) &  ($\mu$s) &  (\%) \\
    \midrule
    \rowcolor{Paired-1!15}                  Radio -      \emph{receive} ($t^\text{Rx}_{1}$)   &      431.83 &    527.32 &  0.94 \\
                                          Stage 1                                             &      431.83 &    527.32 &  0.94 \\ \addlinespace 
                                   Multiplier AGC -    \emph{imultiply} ($t^\text{Rx}_{2}$)   &      367.45 &    619.71 &  1.11 \\
    \rowcolor{Paired-1!15}    Synch. Freq. Coarse -  \emph{synchronize} ($t^\text{Rx}_{3}$)   &      841.32 &    270.66 &  0.48 \\
    \rowcolor{Paired-1!15}         Filter Matched -       \emph{filter} ($t^\text{Rx}_{4}$)   &      116.41 &   1956.08 &  3.49 \\
                                          Stage 2                                             &       80.00 &   2846.45 &  5.08 \\ \addlinespace 
    \rowcolor{Paired-1!15}          Synch. Timing -  \emph{synchronize} ($t^\text{Rx}_{5}$)   &       55.42 &   4108.52 &  7.34 \\
    \rowcolor{Paired-5!15}                Stage 3                                             &       55.42 &   4108.52 &  7.34 \\ \addlinespace 
    \rowcolor{Paired-1!15}          Synch. Timing -      \emph{extract} ($t^\text{Rx}_{6}$)   &      281.83 &    807.97 &  1.44 \\
                                   Multiplier AGC -    \emph{imultiply} ($t^\text{Rx}_{7}$)   &      685.51 &    332.18 &  0.59 \\
    \rowcolor{Paired-1!15}           Synch. Frame -  \emph{synchronize} ($t^\text{Rx}_{8}$)   &      159.41 &   1428.51 &  2.55 \\
                                          Stage 4                                             &       88.65 &   2568.66 &  4.58 \\ \addlinespace 
                                 Scrambler Symbol -   \emph{descramble} ($t^\text{Rx}_{9}$)   &     1682.89 &    135.31 &  0.24 \\
    \rowcolor{Paired-1!15} Synch. Freq. Fine L\&R -  \emph{synchronize} ($t^\text{Rx}_{10}$)  &     1246.85 &    182.63 &  0.33 \\
                            Synch. Freq. Fine P/F -  \emph{synchronize} ($t^\text{Rx}_{11}$)  &      112.56 &   2022.98 &  3.61 \\
                                          Stage 5                                             &       97.27 &   2340.92 &  4.18 \\ \addlinespace 
                                       Framer PLH -       \emph{remove} ($t^\text{Rx}_{12}$)  &     1008.60 &    225.77 &  0.40 \\
                                  Noise Estimator -     \emph{estimate} ($t^\text{Rx}_{13}$)  &      550.06 &    413.98 &  0.74 \\
                                          Stage 6                                             &      355.94 &    639.75 &  1.14 \\ \addlinespace 
                                        Modem PSK -   \emph{demodulate} ($t^\text{Rx}_{14}$)  &       40.47 &   5626.34 & 10.05 \\
                                      Interleaver - \emph{deinterleave} ($t^\text{Rx}_{15}$)  &     1347.25 &    169.02 &  0.30 \\
                                     Decoder LDPC -  \emph{decode SIHO} ($t^\text{Rx}_{16}$)  &      164.21 &   1386.74 &  2.48 \\
                                      Decoder BCH -  \emph{decode HIHO} ($t^\text{Rx}_{17}$)  &        6.92 &  32905.37 & 58.79 \\
                                 Scrambler Binary -   \emph{descramble} ($t^\text{Rx}_{18}$)  &       91.11 &   2499.41 &  4.47 \\
    \rowcolor{Paired-7!15}                Stage 7                                             &        5.35 &  42586.88 & 76.09 \\ \addlinespace 
    \rowcolor{Paired-1!15}       Sink Binary File -         \emph{send} ($t^\text{Rx}_{19}$)  &     1838.31 &    123.87 &  0.22 \\
                                          Stage 8                                             &     1838.31 &    123.87 &  0.22 \\ \midrule      
                                            Total                                             &        4.09 &  55742.37 & 99.57 \\               
    \bottomrule
  \end{tabular}
  }}
\end{table}

Tab.~\ref{tab:sdr_dvbs2_tasks_thr_lat} presents the tasks throughputs and
latencies measured for a sequential execution of the MODCOD 2 in the
transmission phase (x86 target). The tasks have been regrouped per stage in
order to introduce the future decomposition when the parallelism is applied. The
throughputs have been normalized to the number of information bits
($K = 14232$). This enables the comparison among all the reported throughputs.

The stage 7 takes 76\% of the time with especially the \emph{Decoder BCH} task
($t^\text{Rx}_{17}$) that takes 59\% of the time. $t^\text{Rx}_{17}$ should not
take so many time compared to the other tasks. However, we chose to not spend
too much time in optimizing the BCH decoding process as the stage 7 throughput
can easily be increased with the sequence duplication technique. The second
slower stage in the stage 3. This stage is the main hotspot of the implemented
receiver. The stage 3 contains only one synchronization task
($t^\text{Rx}_{5}$). In the current implementation this task cannot be
duplicated (or parallelized) because there is an internal data dependency with
the previous frame (state-full task). The stage 3 is the real limiting factor of
the receiver. If a machine with an infinite number of cores is considered, the
maximum reachable information throughput is 55.42 Mb/s.

We did not try to parallelize the waiting and the learning phases. We measured
that the whole learning phase (1, 2 and 3) takes about one second. During the
learning phase, the receiver is not fast enough to process the received samples
in real time. To fix this problem, the samples are buffered in the \emph{Radio -
receive} task ($t^\text{Rx}_{1}$). Once the learning phase is done, the
transmission phase is parallelized. Thus, the receiver becomes fast enough to
absorb the radio buffer and samples in real time. During the transmission phase,
the receiver is split into 8 stages as presented in
Fig.~\ref{fig:sdr_dvbs2_receiver_transmission}. This decomposition has been
motivated by the nature of the tasks (sequential or parallel) and by the
sequential measured throughput. The number of stages has been minimized in order
to limit the pipeline overhead. Consequently, sequential and parallel tasks
have been regrouped in common stages. The slowest sequential task
($t^\text{Rx}_{5}$) has been isolated in the dedicated stage 3. The other
sequential stages have been formed to always have a higher throughput than the
stage 3. The sequential throughput of the stage 7 (5.35 Mb/s) is lower than the
throughput of the stage 3 (55.42 Mb/s). This is why the sequence duplication has
been applied. The stage 7 has been parallelized over 28 threads. This looks
overkill but the machine was dedicated to the DVB-S2 receiver and the throughput
of the \emph{Decoder LDPC} task ($t^\text{Rx}_{16}$) varies depending on the
SNR. An early termination criterion was enabled. When the
signal quality is very good, the \emph{Decoder LDPC} task runs fast and the
threads can spend a lot of time in waiting. With the passive waiting version of
the adaptor \emph{push} and \emph{pull} tasks, the CPU dynamically adapt the
cores charge and energy can be saved. In Tab.~\ref{tab:sdr_dvbs2_tasks_thr_lat},
the presented \emph{Decoder LDPC} task throughputs and latencies are optimistic
because we are in a SNR error-free zone. All the threads are pinned to a single
core with the \emph{hwloc} library. The 28 threads of the stage 7 are pinned in
round-robin between the CPU sockets. By this way, the memory bandwidth is
maximized thanks to the two NUMA memory banks. The strategy of the stage 7
parallelism is to maximize the throughput. During the duplication process
(modules clones), the thread pinning is known and the memory is copied into the
right memory bank (first touch policy). All the other pipeline stages (1, 2, 3,
4, 5, 6 and 8) are running on a single thread. Because of the synchronizations
between the pipeline stages (adaptor pushes and pulls), the threads have been
pinned on the same socket. The idea is to minimize the pipeline stage latencies
in maximizing the CPU cache performance. It avoids the extra-cost of moving the
cache data between the sockets. On the ARM target, the pipeline has been
decomposed in 12 sequential stages and 1 parallel stage of 40 threads (stage 7).

The receiver program needs around 1.3 GB of the global memory when running in
sequential while it needs around 30 GB in parallel. The memory usage increases
because of the sequence duplications in the stage 7. The duplication operation
takes about 20 seconds. It is made at the very beginning of the program (before
the waiting phase). It is worth mentioning that the amount of memory was not a
critical resource. So, we did not try to reduce its overall occupancy.

\begin{figure}[htp]
  \centering
  \subfloat[][Data copy (stage throughput is 40 Mb/s).\label{plot:sdr_dvbs2_pipeline_copy_dat}]{\includegraphics[width=0.8\linewidth]{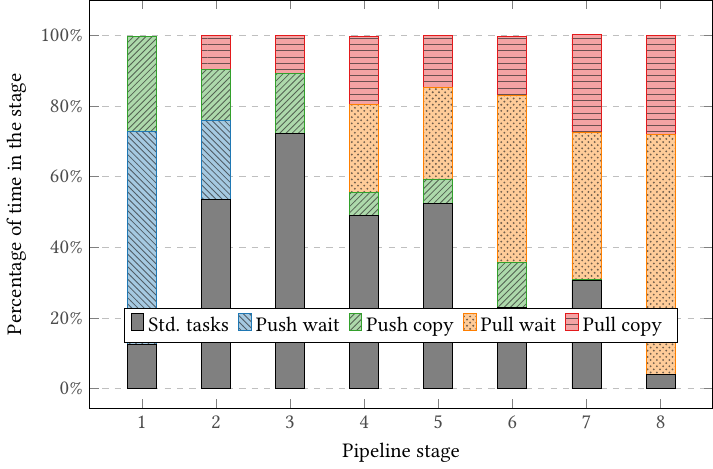}}
  \hfill
  \subfloat[][Copy-less (stage throughput is 55 Mb/s).\label{plot:sdr_dvbs2_pipeline_copy_ptr}]{\includegraphics[width=0.8\linewidth]{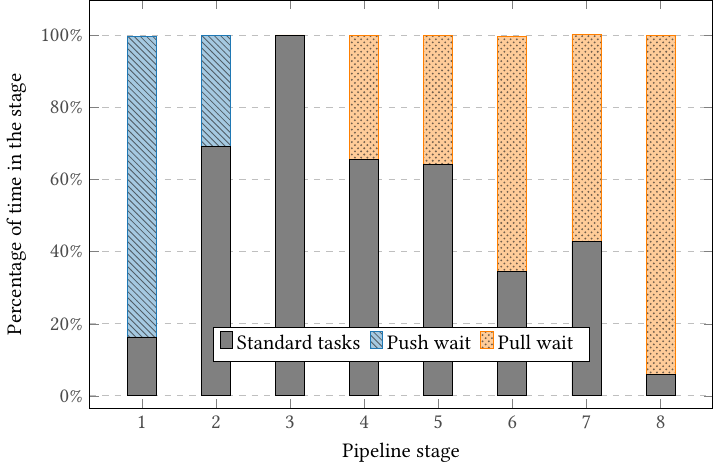}}
  \caption
    [Comparison of the two pipeline implementations in the receiver (x86 target).]
    {Comparison of the two pipeline implementations in the receiver (x86 target, MODCOD 2).}
  \label{plot:sdr_dvbs2_pipeline}
\end{figure}

Fig.~\ref{plot:sdr_dvbs2_pipeline} presents the repartition of the time in the
pipeline stages on the x86 target (MODCOD 2). The receiver is running over 35
threads. Fig.~\ref{plot:sdr_dvbs2_pipeline_copy_dat} shows the pipeline
implementation with data copies. Fig.~\ref{plot:sdr_dvbs2_pipeline_copy_ptr}
shows the pipeline implementation with pointer copies (copy-less). \emph{Push
wait} and \emph{Pull wait} are the percentage of time spent in passive or active
waiting. \emph{Push copy} and \emph{Pull copy} are the percentage of time spent
in copying the data to and from the adaptors buffers. \emph{Standard tasks} is
the cumulative percentage of time spent by the tasks presented in
Fig.~\ref{fig:sdr_dvbs2_receiver_transmission}. In both implementations the
pipeline stage throughput is constraint by the slowest one. In
Fig.~\ref{plot:sdr_dvbs2_pipeline_copy_dat} the measured throughput per stage
is 40 Mb/s whereas in Fig.~\ref{plot:sdr_dvbs2_pipeline_copy_ptr} the measured
throughput is 55 Mb/s. The copy-less implementation throughput is $\approx$ 27\%
higher than the data copy implementation.
Fig.~\ref{plot:sdr_dvbs2_pipeline_copy_dat} shows that the copy overhead is
non-negligible. A 27\% slowdown is directly due to these copies in the stage 3.
It largely justifies the copy-less implementation. In
Fig.~\ref{plot:sdr_dvbs2_pipeline_copy_ptr} and in the stage 3, 100\% of time
is taken by the $t^\text{Rx}_{5}$ task. This is also confirmed by the measured
throughput (55 Mb/s) which is very close the sequential throughput (55.42
Mb/s) reported in Tab.~\ref{tab:sdr_dvbs2_tasks_thr_lat}.

\begin{table}[t]
  \centering
  \caption
    {Throughputs depending on the selected DVB-S2 configuration.}
  \label{tab:sdr_dvbs2_thr_modcod}
  \begin{tabular}{c c c c c c c}
    \toprule
                         & \multicolumn{4}{c}{\text{Throughput} (Mb/s)} & & \\ \cmidrule(lr){2-5}
                         & \multicolumn{2}{c}{\text{Sequential}} & \multicolumn{2}{c}{\text{Parallel}} & \multicolumn{2}{c}{Latency (ms)} \\ \cmidrule(lr){2-3} \cmidrule(lr){4-5} \cmidrule(lr){6-7}
    \text{Configuration} & \text{x86} & \text{ARM} & \text{x86} & \text{ARM} & \text{x86} & \text{ARM} \\
    \midrule
    MODCOD 1 & 3.4 & 1.0 & 37 & 19 & -- & 37 \\
    MODCOD 2 & 4.1 & 1.4 & 55 & 28 & 56 & 41 \\
    MODCOD 3 & 4.0 & 1.1 & 80 & 42 & -- & 51 \\
    \bottomrule
  \end{tabular}
\end{table}

Tab.~\ref{tab:sdr_dvbs2_thr_modcod} summarizes sequential and parallel
throughputs for the 3 MODCODs presented in Tab.~\ref{tab:sdr_dvbs2_modcod}. To
measure the maximum achievable throughput, the USRP modules are removed and
samples are read from a binary file. This is because the pipeline stages are
naturally adapting to the slowest one. It means that in a real communication,
the throughput of the radio is always configured to be just a little bit slower
than the slowest stage. Otherwise the radio task has to indefinitely buffer
samples even though the amount of available memory in the machine is not
infinite. The information throughput ($K$ bits) is the final useful throughput
for the user. Between the MODCOD 1 and 2, only the LDPC code rate varies
($R=3/5$ and $R=8/9$ resp.). In the parallel implementation, it has a direct
impact on the information throughput. Between the MODCOD 2 and 3, the modulation
varies (QPSK and 8-PSK resp.) and the frames have to be deinterleaved
(column/row interleaver). High order modulation reduces the amount of samples
processed in the \emph{Synchronizer Timing} task ($t^\text{Rx}_{5}$): this
results in higher throughput (80 Mb/s for the 8-PSK) in the slowest stage~3. In
the parallel implementation, the pipeline stage throughputs are adapting to the
slowest stage 3. It results in an important speedup. In the sequential
implementation, it results in a little slowdown. Indeed, the additional time
spent in the \emph{deinterleave} task ($t^\text{Rx}_{15}$) is higher than the
time saved in the \emph{Synchronizer Timing} task ($t^\text{Rx}_{5}$).

These results demonstrate the benefit of our parallel implementation.
Throughput speedups range from 10 to 20 compared to the sequential
implementation. Selected configurations each are most efficient in different SNR
zones (as shown in Fig.~\ref{fig:sdr_dvbs2_bfer}), depending on the signal
quality. For instance, MODCOD~1 is adapted for noisy environments (3 dB).
However the information throughput is limited to 37 Mb/s (x86 target). MODCOD~3
is more adapted to clearer signal conditions (7.5 dB) and the information
throughput reaches 80~Mb/s (x86 target). MODCOD~2 is in-between. The throughputs
obtained on the ARM target are lower than on the x86 CPUs (by a factor of
$\approx$ 2 when running in parallel). It can be explained by the limited
mono-core performance of the ThunderX2 architecture: the frequency is lower
(2.0 GHz versus 2.7 GHz) and the SIMD width is smaller (128-bit in NEON versus
512-bit in AVX-512F). However, being able to run the transceiver on both x86 and
ARM CPUs with comparable throughput demonstrates the flexibility and the
portability of the proposed framework.

\subsection{Comparison with State-of-the-Art}
\label{sec:comparison}

\paragraph{gr-dvbs2rx} As we said before, to the best of our knowledge, it is
the faster open source implementation at the time of the writing of the paper.
gr-dvbs2rx has been run on the same x86 target presented before (with the same
compiler and options) and on the MODCOD 2. We ran the code without the radios,
this way only the software part of the receiver is evaluated. First, a set of
IQs have been generated from the emitter and written on a file. Then, the
receiver has been executed on it. The set of IQs have been read from the same
file. The evaluation has been made on error-free SNR zone. To make a fair
comparison, we modified a little bit the source code of the receiver to remove
``artificial blocks'' that slowed down the throughput. The throttle block has
been removed as well as some useless (and not optimized) blocks dedicated to the
conversion of the IQs (from 8-bit fixed-point to 32-bit floating-point format).
The source code modifications we made are available on a fork of the
project\footnote{gr-dvbs2rx fork (thr\_benchmark branch):
\url{github.com/kouchy/gr-dvbs2rx/}}.

\begin{table}[htp]
  \centering
  \caption
    {gr-dvbs2rx throughputs per pipeline stage compared with this work
     (error-free SNR zone, x86 target, MODCOD 2). Same color codes as in
     Tab.~\ref{tab:sdr_dvbs2_tasks_thr_lat}.}
  \label{tab:sdr_dvbs2_gr_thr}
  {\resizebox{1.0\linewidth}{!}{
  \begin{tabular}{r r r r r}
    \toprule
    \multicolumn{2}{c}{GNU Radio} & Equi. & \multicolumn{2}{c}{\text{Throughput} (Mb/s)} \\ \cmidrule(lr){1-2} \cmidrule(lr){4-5}
    Stage & Block & ($t^\text{Rx}_{\{i\}}$) & gr-dvbs2rx & This work \\
    \midrule
    \rowcolor{Paired-1!15}  1 & \textit{file\_source}      & 1      & 100.7 &  431.8 \\
                            2 & \textit{agc\_cc}           & 2      &  24.0 &  367.5 \\
    \rowcolor{Paired-5!15}  3 & \textit{symbol\_sync\_cc}  & 4-6    &  16.9 &   33.1 \\
                            4 & \textit{rotator\_cc}       & 7      &  96.1 &  685.5 \\
    \rowcolor{Paired-1!15}  5 & \textit{plsync\_cc}        & 3,8-13 &  45.3 &   48.4 \\
                            6 & \textit{ldpc\_decoder\_cb} & 14-16  &  23.0 &   31.7 \\
    \rowcolor{Paired-7!15}  7 & \textit{bch\_decoder\_bb}  & 17     &  14.9 &    6.9 \\
                            8 & \textit{bbdescrambler\_bb} & 18     & 225.8 &   91.1 \\
                            9 & \textit{bbdeheader\_bb}    & --     & 252.5 &     -- \\
    \rowcolor{Paired-1!15} 10 & \textit{file\_sink}        & 19     & 346.5 & 1838.3 \\
    \bottomrule
  \end{tabular}
  }}
\end{table}

Tab.~\ref{tab:sdr_dvbs2_gr_thr} presents the per block normalized throughputs
of gr-dvbs2rx and of this work (considering the gr-dvbs2rx pipeline
decomposition). For each GNU Radio block the tasks equivalence with our system
is given. We measured an overall information throughput of 14.9 Mb/s. As GNU
Radio pins each block to a thread, the throughput performance is driven by the
slowest block (here \textit{bch\_decoder\_bb}). gr-dvbs2rx uses 10 threads (same
as the number of stages). If we applied the same decomposition without
the duplication of the stage 7 our receiver will have been limited to the
throughput of the BCH decoder (6.9 Mb/s). With the duplication technique the
stage 7 is automatically dispatched on multiple threads and its throughput is
approximately multiplied by the number of threads. This automatic transformation
is not possible with the GNU Radio implementation. Moreover, still if we only
consider the gr-dvbs2rx pipeline decomposition, our work will have been limited
to the lowest sequential throughput which is 33.1 Mb/s
(\textit{symbol\_sync\_cc} block). Thanks to a finer decomposition into tasks
($\text{\textit{symbol\_sync\_cc}} = t^\text{Rx}_{4} + t^\text{Rx}_{5} +
t^\text{Rx}_{6}$) our receiver is able to reach 55 Mb/s (see
Tab.~\ref{tab:sdr_dvbs2_thr_modcod}). As a result, the throughput of the
proposed implementation is 3.7 times faster than gr-dvbs2rx.

\paragraph{Grayver and Utter} On a comparable CPU, we estimated that their work
is able to double or even triple the throughput of our implementation. This is
mainly due to new algorithmic improvements in the synchronization tasks. For
instance, they were able to express more parallelism in the \emph{Synchronizer
Timing} task ($t^\text{Rx}_{5}$). However, we also tried some aggressive
optimizations in this task but we never succeeded to measure the same level of
FER decoding performance. It could be interesting to check for any penalty in
terms of decoding performance that may occur and to combine their optimizations
with our DSEL. Unlike our work, their work focuses on a single DVB-S2
configuration (8-PSK, $N = 64800$ and $R = 1/2$) and a single architecture
(x86). Their implementation looks like an hard-coded solution for the DVB-S2
standard while our goal is to provide generic methods and tools for SDR system
implementations.

\section{Conclusion}
\label{sec:conclu}

In this article, we introduced a new DSEL designed to satisfy SDR needs in terms
of expressiveness and performance on multicore processors. It allows the
definition of stateful tasks, early exits and dynamic control. We evaluated it
on micro-benchmarks and showed that its scheduling overhead is negligible for
tasks longer than 4~$\mu$s. We evaluated a full software implementation of the
DVB-S2 standard built with our DSEL and the \AFFECT library for tasks. This
implementation is the fastest open source software solution on multicore CPUs.
It matches satellite real time constraints (30 - 50 Mb/s), which demonstrates
the relevance and efficiency of the DSEL. This is the consequence of two main
factors: 1) the low overhead achieved by DSEL, 2) an efficient implementation of
the pipeline technique, where one stage, parallelized, reaches saturation. In
future works, we plan to integrate the parallel features of the DSEL with high
level languages such as Python or MATLAB\R typically used in the signal
processing community, often less familiar with the \Cxx language. Moreover,
automatic parallelization, tuning of pipelining stages, could be investigated. A
profile-guided optimization, capturing task runtime, has been used to tune
pipeline stages, for instance. A more automatic and integrated approach could be
possible since the analysis of the task graph is dynamic.

\bibliographystyle{IEEEtran}
\bibliography{refs.bib}

\end{document}